\documentclass[11pt,a4paper]{article}
\usepackage[hyperref]{acl2020}
\usepackage{float}
\usepackage{graphicx}
\usepackage{times}
\usepackage{latexsym}
\usepackage{listings}
\usepackage{color}

\usepackage{microtype}

\aclfinalcopy 

\definecolor{mygreen}{rgb}{0,0.6,0}
\definecolor{mygray}{rgb}{0.5,0.5,0.5}
\definecolor{mymauve}{rgb}{0.58,0,0.82}

\title{\texttt{jiant}: A Software Toolkit for Research \\on General-Purpose Text Understanding Models}

\author{Yada Pruksachatkun,\textsuperscript{1}\thanks{~~Equal contribution.}\textsuperscript{ } Phil Yeres,\textsuperscript{1}\footnotemark[1]\textsuperscript{ } Haokun Liu,\textsuperscript{1} Jason Phang,\textsuperscript{1} \\ 
  \bf Phu Mon Htut,\textsuperscript{1} Alex Wang,\textsuperscript{1} Ian Tenney\textsuperscript{2}, Samuel R. Bowman\textsuperscript{1} \\
  \textsuperscript{1}New York University, \textsuperscript{2}Google Research \\
  \texttt{\{yp913,bowman\}@nyu.edu}}

\date{}

\begin{document}
\maketitle
\begin{abstract}
We introduce \texttt{jiant}, an open source toolkit for conducting multitask and transfer learning experiments on English NLU tasks. \texttt{jiant} enables modular and configuration-driven experimentation with state-of-the-art models and implements a broad set of tasks for probing, transfer learning, and multitask training experiments. \texttt{jiant} implements over 50 NLU tasks, including all GLUE and SuperGLUE benchmark tasks. We demonstrate that \texttt{jiant} reproduces published performance on a variety of tasks and models, including BERT and RoBERTa. \texttt{jiant} is available at \url{https://jiant.info}.
\end{abstract}

\section{Introduction}
This paper introduces \texttt{jiant},\footnote{The name \texttt{jiant} stands for ``jiant is an NLP toolkit".} an open source toolkit that allows researchers to quickly experiment on a wide array of NLU tasks, using state-of-the-art NLP models, and conduct experiments on probing, transfer learning, and multitask training.  \texttt{jiant} supports many state-of-the-art Transformer-based models implemented by Huggingface's Transformers package, as well as non-Transformer models such as BiLSTMs.

Packages and libraries like HuggingFace's Transformers \citep{Wolf2019HuggingFacesTS} and AllenNLP \citep{Gardner2017AllenNLP} have accelerated the process of experimenting and iterating on NLP models by both abstracting out implementation details, and simplifying the model training pipeline. 
\texttt{jiant} extends the capabilities of both toolkits by presenting a wrapper that implements a variety of complex experimental pipelines in a scalable and easily controllable setting. 
\texttt{jiant} contains a task bank of over 50 tasks, including all the tasks presented in GLUE \citep{wang2018glue}, SuperGLUE \citep{wang2019superglue}, the edge-probing suite \cite{tenney2018what}, and the SentEval probing suite \citep{conneau-kiela-2018-senteval}, as well as other individual tasks including CCG supertagging \citep{ccg}, SocialIQA \citep{sap2019socialiqa}, and CommonsenseQA \citep{talmor2018commonsenseqa}.
\texttt{jiant} is also the official baseline codebase for the SuperGLUE benchmark.

\texttt{jiant}'s core design principles are:
\begin{itemize}
    \item Ease of use: \texttt{jiant} should allow users to run a variety of experiments using state-of-the-art models via an easy to use configuration-driven interface. 
    \item Reproducibility: \texttt{jiant} should provide features that support correct and reproducible experiments, including logging and saving and restoring model state.
    \item Availability of NLU tasks: \texttt{jiant} should maintain and continue to grow a collection of tasks useful for NLU research, especially popular evaluation tasks and tasks commonly used in pretraining and transfer learning.
    \item Availability of cutting-edge models: \texttt{jiant} should make implementations of state-of-the-art models available for experimentation.
    \item Open source: \texttt{jiant} should be free to use, and easy to contribute to.
\end{itemize}

Early versions of \texttt{jiant} have already been used in multiple works, including probing analyses \citep{tenney2018what,tenney2019bert,warstadt-etal-2019-investigating,lin-etal-2019-open,hewitt2019structural,jawahar-etal-2019-bert}, transfer learning experiments \citep{wang2018tell,Phang2018SentenceEO}, and dataset and benchmark construction \citep{wang2019superglue, wang2018glue, functionalprobe}.

 \begin{figure}
  \includegraphics[width=\columnwidth]{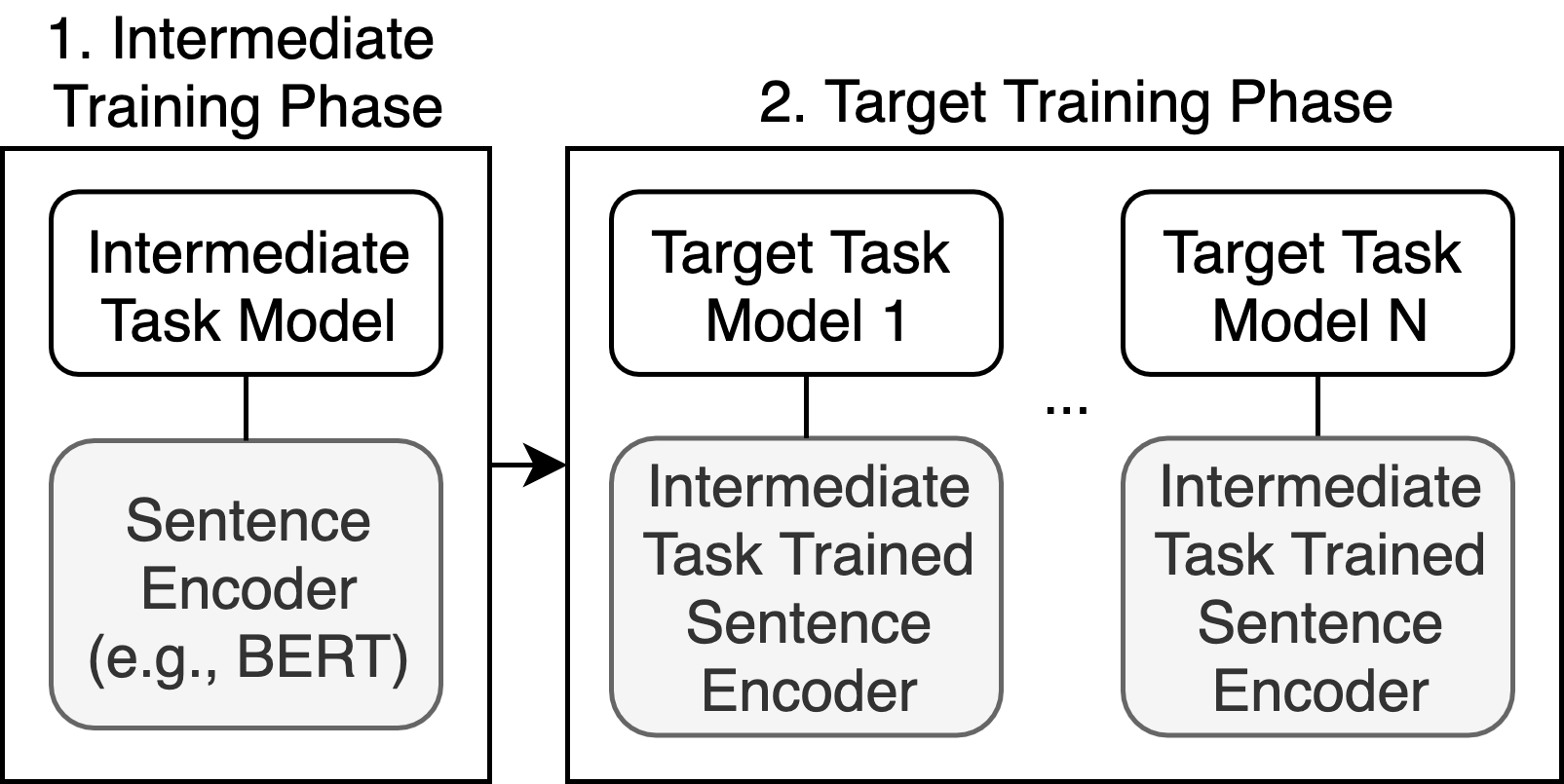}
  \caption{Multi-phase \texttt{jiant} experiment configuration used by \citet{wang2018tell}: a BERT sentence encoder is trained with an intermediate task model during \texttt{jiant}'s intermediate training phase, and fine-tuned with various target task models in \texttt{jiant}'s target training phase.}
  \label{fig:edge}
\end{figure}

\section{Background}
Transfer learning is an area of research that uses knowledge from pretrained models to transfer to new tasks. In recent years, Transformer-based models like BERT \citep{devlin-etal-2019-bert} and T5 \citep{2019t5} have yielded state-of-the-art results on the lion's share of benchmark tasks for language understanding through pretraining and transfer, often paired with some form of multitask learning.

\texttt{jiant} enables a variety of complex training pipelines through simple configuration changes, including multi-task training \citep{Caruana1993MultitaskLA,liu-etal-2019-multi} and pretraining, as well as the sequential fine-tuning approach from STILTs \citep{Phang2018SentenceEO}. 
In STILTs, intermediate task training takes a pretrained model like ELMo or BERT, and applies supplementary training on a set of intermediate tasks, before finally performing single-task training on additional downstream tasks.

\section{\texttt{jiant} System Overview}

\subsection{Requirements and Deployment}
\texttt{jiant} can be cloned and installed from GitHub: \url{https://github.com/nyu-mll/jiant}. \texttt{jiant} v1.3.0 requires Python 3.5 or later, and \texttt{jiant}'s core dependencies are PyTorch \citep{NEURIPS2019_9015}, AllenNLP \citep{Gardner2017AllenNLP}, and HuggingFace's Transformers \citep{Wolf2019HuggingFacesTS}. \texttt{jiant} is released under the MIT License \citep{osi2020}.
 \texttt{jiant} runs on consumer-grade hardware or in cluster environments with or without CUDA GPUs. The \texttt{jiant} repository also contains documentation and configuration files demonstrating how to deploy \texttt{jiant} in Kubernetes clusters on Google Kubernetes Engine.

\subsection{\texttt{jiant} Components}
\begin{figure*}
  \includegraphics[width=\textwidth]{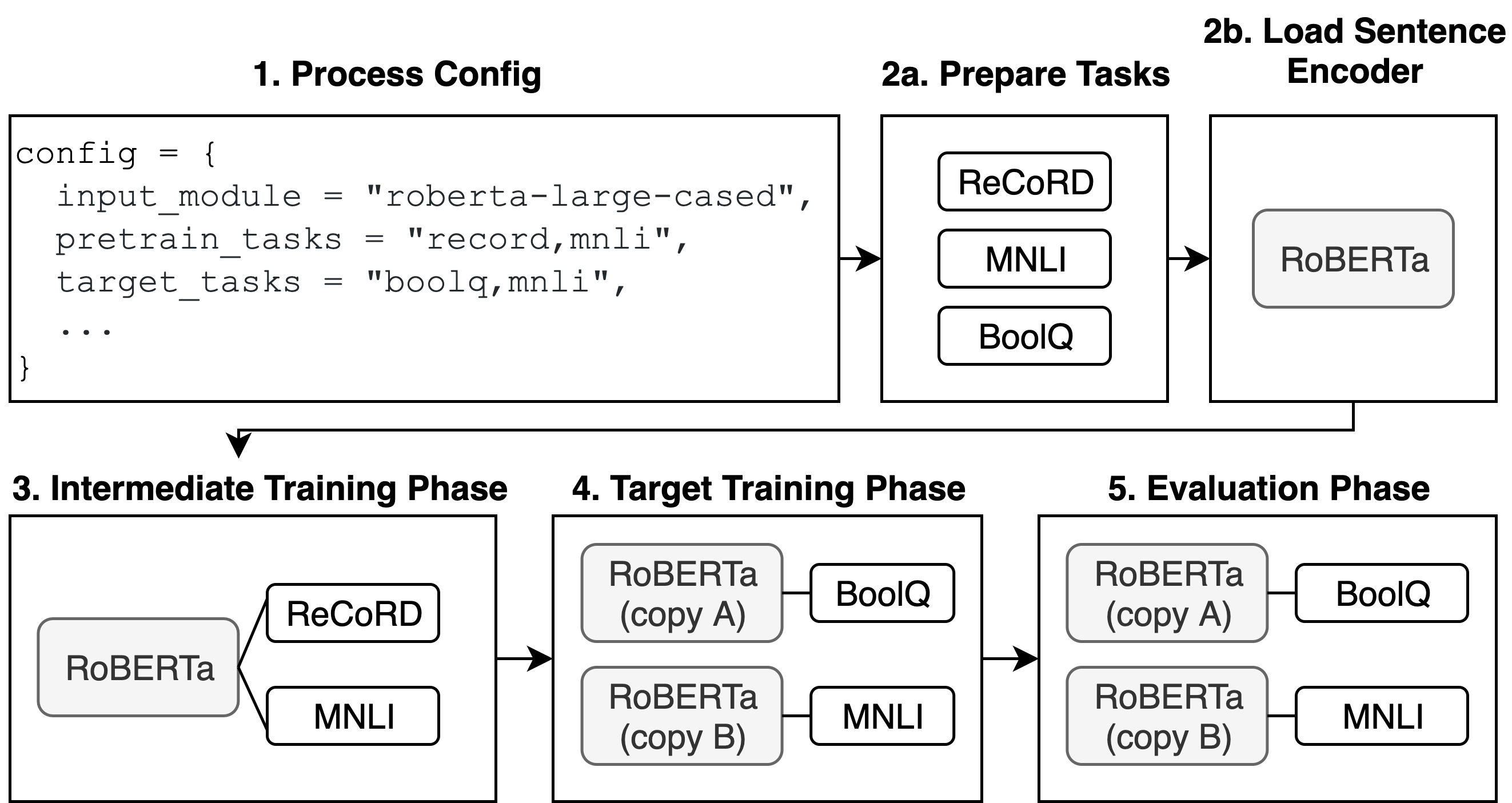}
  \caption{\texttt{jiant} pipeline stages using RoBERTa as the sentence encoder, ReCoRD and MNLI tasks as intermediate tasks, and MNLI and BoolQ as tasks for target training and evaluation. The diagram highlights that during target training and evaluation phases, copies are made of the sentence encoder model, and fine tuning and evaluation for each task are conducted on separate copies.}
  \label{fig:flow}
\end{figure*}

\begin{itemize}
\item Tasks: Tasks have references to task data, methods for processing data, references to classifier heads, and methods for calculating performance metrics, and making predictions.
\item Sentence Encoder: Sentence encoders map from the indexed examples to a sentence-level representation. Sentence encoders can include an input module (e.g., Transformer models, ELMo, or word embeddings), followed by an optional second layer of encoding (usually a BiLSTM). Examples of possible sentence encoder configurations include BERT, ELMo followed by a BiLSTM, BERT with a variety of pooling and aggregation methods, or a bag of words model.
\item Task-Specific Output Heads: Task-specific output modules map representations from sentence encoders to outputs specific to a task, e.g. entailment/neutral/contradiction for NLI tasks, or tags for part-of-speech tagging. They also include logic for computing the corresponding loss for training (e.g. cross-entropy).
\item Trainer: Trainers manage the control flow for the training and validation loop for experiments. They sample batches from one or more tasks, perform forward and backward passes, calculate training metrics, evaluate on a validation set, and save checkpoints. Users can specify experiment-specific parameters such as learning rate, batch size, and more. 
\item Config: Config files or flags are defined in HOCON\footnote{Human-Optimized Config Object Notation \citep{lightbend2011}. \texttt{jiant} uses HOCON's logic to consolidate multiple config files and command-line overrides into a single run config.} format. Configs specify parameters for \texttt{jiant} experiments including choices of tasks, sentence encoder, and training routine.\footnote{\texttt{jiant} configs support multi-phase training routines as described in section \ref{pipeline} and illustrated in Figure \ref{fig:flow}.}
\end{itemize}

Configs are \texttt{jiant}'s primary user interface. Tasks and modeling components are designed to be modular, while \texttt{jiant}'s pipeline is a monolithic, configuration-driven design intended to facilitate a number of common workflows outlined in \ref{pipeline}.

\subsection{\texttt{jiant} Pipeline Overview} \label{pipeline}
\texttt{jiant}'s core pipeline consists of the five stages described below and illustrated in Figure \ref{fig:flow}:
\begin{enumerate}
  \item A config or multiple configs defining an experiment are interpreted. Users can choose and configure models, tasks, and stages of training and evaluation.
  \item The tasks and sentence encoder are prepared:
  
  \begin{enumerate}
    \item The task data is loaded, tokenized, and indexed, and the preprocessed task objects are serialized and cached. In this process, AllenNLP is used to create the vocabulary and index the tokenized data. 
    \item The sentence encoder is constructed and (optionally) pretrained weights are loaded.\footnote{ The sentence encoder's weights can optionally be left frozen, or be included in the training procedure.}
    \item The task-specific output heads are created for each task, and task heads are attached to a common sentence encoder. Optionally, different tasks can share the same output head,  as in \citet{liu-etal-2019-multi}.
  \end{enumerate}
  
  \item Optionally, in the intermediate phase the trainer samples batches randomly from one or more tasks,\footnote{ Tasks can be sampled using a variety of sample weighting methods, e.g., uniform or proportional to the tasks' number of training batches or examples.} and trains the shared model.
  \item Optionally, in the target training phase, a copy of the model is configured and trained or fine-tuned for each target task separately.
  \item Optionally, the model is evaluated on the validation and/or test sets of the target tasks.
\end{enumerate}

\subsection{Task and Model resources in \texttt{jiant}}
\texttt{jiant} supports over 50 tasks. Task types include classification, regression, sequence generation, tagging, masked language modeling, and span prediction. \texttt{jiant} focuses on NLU tasks like MNLI \citep{N18-1101}, CommonsenseQA \citep{talmor2018commonsenseqa}, the Winograd Schema Challenge \citep{wsc}, and SQuAD \cite{squad}. A full inventory of tasks and task variants is available in the \href{https://github.com/nyu-mll/jiant/tree/master/jiant/tasks}{\texttt{jiant/tasks}} module.

\texttt{jiant} provides support for cutting-edge sentence encoder models, including support for Huggingface's Transformers. Supported models include:
ELMo \citep{peters-etal-2018-deep}, GPT \citep{radford2018improving}, BERT \citep{devlin-etal-2019-bert}, XLM \citep{NIPS20198928}, GPT-2 \citep{radford2019language}, XLNet \citep{yang2019xlnet}, RoBERTa \citep{liu2019roberta}, and ALBERT \citep{lan2019albert}.
\texttt{jiant} also supports the from-scratch training of (bidirectional) LSTMs \citep{hochreiter1997long} and deep bag of words models \citep{iyyer-etal-2015-deep}, as well as syntax-aware models such as PRPN \cite{DBLP:conf/iclr/ShenLHC18} and ON-LSTM \cite{shen2018ordered}. \texttt{jiant} also supports  word embeddings such as GloVe \citep{pennington-etal-2014-GloVe}.

\begin{figure}[t]
\begin{lstlisting}[language=java,frame = single,xleftmargin=.025\textwidth,xrightmargin=.025\textwidth,]
// Config for BERT experiments.

// Get default configs from a file:
include "defaults.conf"
exp_name = "bert-large-cased"

// Data and preprocessing settings
max_seq_len = 256 

// Model settings
input_module = "bert-large-cased"
transformers_output_mode = "top"
s2s = {
    attention = none
}
sent_enc = "none"
sep_embs_for_skip = 1
classifier = log_reg 
// fine-tune entire BERT model
transfer_paradigm = finetune

// Training settings
dropout = 0.1
optimizer = bert_adam
batch_size = 4
max_epochs = 10
lr = .00001
min_lr = .0000001
lr_patience = 4
patience = 20
max_vals = 10000

// Phase configuration
do_pretrain = 1
do_target_task_training = 1
do_full_eval = 1
write_preds = "val,test"
write_strict_glue_format = 1

// Task specific configuration 
commitbank = {
    val_interval = 60
    max_epochs = 40
}
\end{lstlisting}
\caption{Example \texttt{jiant} experiment config file.}\label{tab:config}
\end{figure}

\subsection{User Interface}
\texttt{jiant} experiments can be run with a simple CLI:
\begin{lstlisting}[basicstyle=\ttfamily\footnotesize,]
python -m jiant \
  --config_file roberta_with_mnli.conf \
  --overrides "target_tasks = swag, \
               run_name = swag_01"
\end{lstlisting}

\texttt{jiant} provides default config files that allow running many experiments without modifying source code.

\texttt{jiant} also provides baseline config files that can serve as a starting point for model development and evaluation against GLUE \citep{wang2018glue} and SuperGLUE \citep{wang2019superglue} benchmarks.

More advanced configurations can be developed by composing multiple configurations files and overrides. Figure \ref{tab:config} shows a config file that overrides a default config, defining an experiment that uses BERT as the sentence encoder. This config includes an example of a task-specific configuration, which can be overridden in another config file or via a command line override.

Because \texttt{jiant} implements the option to provide command line overrides with a flag, it is easy to write scripts that launch \texttt{jiant} experiments over a range of parameters, for example while performing grid search across hyperparameters. \texttt{jiant} users have successfully run large-scale experiments launching hundreds of runs on both Kubernetes and Slurm.

\subsection{Example \texttt{jiant} Use Cases and Options}
Here we highlight some example use cases and key corresponding \texttt{jiant} config options required in these experiments:
\begin{itemize}
\itemsep-0.3em
    \item Fine-tune BERT on SWAG \citep{zellers-etal-2018-swag} and SQUAD \cite{squad}, then fine-tune on HellaSwag \citep{zellers-etal-2019-hellaswag}:
        \begin{lstlisting}
input_module = bert-base-cased
pretrain_tasks = "swag,squad"
target_tasks = hellaswag
        \end{lstlisting}
           \item Train a probing classifier over a frozen BERT model, as in \citet{tenney2019bert}:
        \begin{lstlisting}
input_module = bert-base-cased
target_tasks = edges-dpr
transfer_paradigm = frozen
        \end{lstlisting}
        
    \item Compare performance of GloVe \citep{pennington-etal-2014-GloVe} embeddings using a BiLSTM:
        \begin{lstlisting}
input_module = glove
sent_enc = rnn
        \end{lstlisting}
    \item Evaluate ALBERT \citep{lan2019albert} on the MNLI \citep{N18-1101} task:
        \begin{lstlisting}
input_module = albert-large-v2
target_task = mnli
        \end{lstlisting}
\end{itemize}

\subsection{Optimizations and Other Features}
\texttt{jiant} implements features that improve run stability and efficiency:
\begin{itemize}
  \item \texttt{jiant} implements checkpointing options designed to offer efficient early stopping and to show consistent behavior when restarting after an interruption.
  \item \texttt{jiant} caches preprocessed task data to speed up reuse across experiments which share common data resources and artifacts.
  \item \texttt{jiant} implements gradient accumulation and multi-GPU, which enables training on larger batches than can fit in memory for a single GPU.
   \item \texttt{jiant} supports outputting predictions in a format ready for GLUE and SuperGLUE benchmark submission.
   \item \texttt{jiant} generates custom log files that capture experimental configurations, training and evaluation metrics, and relevant run-time information. 
   \item \texttt{jiant} generates TensorBoard event files \citep{tensorflow2015-whitepaper} for training and evaluation metric tracking. TensorBoard event files can be visualized using the TensorBoard Scalars Dashboard.
\end{itemize}

\subsection{Extensibility}
\texttt{jiant}'s design offers conveniences that reduce the need to modify code when making changes:
\begin{itemize}
    \item \texttt{jiant}'s task registry makes it easy to define a new version of an existing task using different data. Once the new task is defined in the task registry, the task is available as an option in \texttt{jiant}'s config.
  \item \texttt{jiant}'s sentence encoder and task output head abstractions allow for easy support of new sentence encoders.
\end{itemize}

In use cases requiring the introduction of a new task, users can use class inheritance to build on a number of available parent task types including classification, tagging, span prediction, span classification, sequence generation, regression, ranking, and multiple choice task classes. For these task types, corresponding task-specific output heads are already implemented.

More than 30 researchers and developers from more than 5 institutions have contributed code to the \texttt{jiant} project.\footnote{\url{https://github.com/nyu-mll/jiant/graphs/contributors}}
\texttt{jiant}'s maintainers welcome pull requests that introduce new tasks or sentence encoder components, and pull request are actively reviewed.
The \texttt{jiant} repository's continuous integration system requires that all pull requests pass unit and integration tests and meet Black\footnote{\url{https://github.com/psf/black}} code formatting requirements.

\subsection{Limitations and Development Roadmap}
While \texttt{jiant} is quite flexible in the pipelines that can be specified through configs, and some components are highly modular (e.g., tasks, sentence encoders, and output heads), modification of the pipeline code can be difficult. For example, training in more than two phases would require modifying the trainer code.\footnote{While not supported by config options, training in more than two phases is possible by using \texttt{jiant}'s checkpointing features to reload models for additional rounds of training.} Making multi-stage training configurations more flexible is on \texttt{jiant}'s development roadmap.

\texttt{jiant}'s development roadmap prioritizes adding support for new Transformer models, and adding tasks that are commonly used for pretraining and evaluation in NLU. Additionally, there are plans to make \texttt{jiant}'s training phase configuration options more flexible to allow training in more than two phases, and to continue to refactor \texttt{jiant}'s code to keep \texttt{jiant} flexible to track developments in NLU research. 

\section{Benchmark Experiments}
To benchmark \texttt{jiant}, we perform a set of experiments that reproduce external results for single fine-tuning and transfer learning experiments. \texttt{jiant} has been benchmarked extensively in both published and ongoing work on a majority of the implemented tasks.

We benchmark single-task fine-tuning configurations using CommonsenseQA \cite{talmor2018commonsenseqa} and SocialIQA \cite{sap2019socialiqa}.
On CommonsenseQA with $\mathrm{RoBERTa}_\mathrm{LARGE}$, \texttt{jiant} achieves an accuracy of 72.2, comparable to 72.1 reported by \citet{liu2019roberta}. On SocialIQA with BERT-large, \texttt{jiant} achieves a dev set accuracy of 65.8, comparable to 66.0 reported in \citet{sap2019socialiqa}.

Next, we benchmark \texttt{jiant}'s transfer learning regime. We perform transfer experiments from MNLI to BoolQ with BERT-large. In this configuration \citet{clark-etal-2019-boolq} demonstrated an accuracy improvement of 78.1 to 82.2 on the dev set, and \texttt{jiant} achieves an improvement of 78.1 to 80.3.

\section{Conclusion}

\texttt{jiant} provides a configuration-driven interface for defining transfer learning and representation learning experiments using a bank of over 50 NLU tasks, cutting-edge sentence encoder models, and multi-task and multi-stage training procedures. Further, \texttt{jiant} is shown to be able to replicate published performance on various NLU tasks.

\texttt{jiant}'s modular design of task and sentence encoder components make it possible for users to quickly and easily experiment with a large number of tasks, models, and parameter configurations, without editing source code. \texttt{jiant}'s design also makes it easy to add new tasks, and \texttt{jiant}'s architecture makes it convenient to extend \texttt{jiant} to support new sentence encoders.

\texttt{jiant} code is open source, and \texttt{jiant} invites contributors to open issues or submit pull request to the \texttt{jiant} project repository: \url{https://github.com/nyu-mll/jiant}.

\section*{Acknowledgments}
Katherin Yu, Jan Hula, Patrick Xia, Raghu Pappagari, Shuning Jin, R. Thomas McCoy, Roma Patel, Yinghui Huang, Edouard Grave, Najoung Kim, Thibault F\'evry, Berlin Chen, Nikita Nangia, Anhad Mohananey, Katharina Kann, Shikha Bordia, Nicolas Patry, David Benton, and Ellie Pavlick have contributed substantial engineering assistance to the project.

The early development of \texttt{jiant} took at the 2018 Frederick Jelinek Memorial Summer Workshop on Speech and Language Technologies, and was supported by Johns Hopkins University with unrestricted gifts from Amazon, Facebook, Google, Microsoft and Mitsubishi Electric Research Laboratories. 

Subsequent development was possible in part by a donation to NYU from Eric and Wendy Schmidt made by recommendation of the Schmidt Futures program, by support from Intuit Inc., and by support from Samsung Research under the project \textit{Improving Deep Learning using Latent Structure}. We gratefully acknowledge the support of NVIDIA Corporation with the donation of a Titan V GPU used at NYU in this work. Alex Wang's work on the project is supported by the National Science Foundation Graduate Research Fellowship Program under Grant No. DGE 1342536. Any opinions, findings, and conclusions or recommendations expressed in this material are those of the author(s) and do not necessarily reflect the views of the National Science Foundation. Yada Pruksachatkun's work on the project is supported in part by the Moore-Sloan Data Science Environment as part of the NYU Data Science Services initiative. Sam Bowman's work on \texttt{jiant} during Summer 2019 took place in his capacity as a visiting researcher at Google.

\bibliography{acl2020}
\bibliographystyle{acl_natbib}

\clearpage

\appendix

\end{document}